\documentclass{article}
\usepackage[utf8]{inputenc}

\usepackage[letterpaper,top=2cm,bottom=2cm,left=3cm,right=3cm,marginparwidth=1.75cm]{geometry}

\usepackage{amsmath}
\usepackage{graphicx}

\usepackage{hyperref}

\usepackage[sort&compress,numbers,square]{natbib}

\hypersetup{pdftitle={Multi--agent Interplay in a Competitive Survival Environment},pdfauthor={Andrea Fanti}}

\begin{document}
    \title{Multi-Agent Interplay in a Competitive Survival Environment}

\author{
Andrea Fanti\\Sapienza University of Rome
}

\date{}

\maketitle

\begin{abstract}
Solving hard--exploration environments in an important challenge in Reinforcement Learning. Several approaches have been proposed and studied, such as Intrinsic Motivation, co--evolution of agents and tasks, and multi--agent competition. In particular, the interplay between multiple agents has proven to be capable of generating human--relevant emergent behaviour that would be difficult or impossible to learn in single--agent settings. In this work, an extensible competitive environment for multi--agent interplay was developed, which features realistic physics and human--relevant semantics. Moreover, several experiments on different variants of this environment were performed, resulting in some simple emergent strategies and concrete directions for future improvement. The content presented here is part of the author's thesis ``Multi--Agent Interplay in a Competitive Survival Environment'' for the Master's Degree in Artificial Intelligence and Robotics at Sapienza University of Rome, 2022.
\end{abstract}

    \section{Introduction}

Creating artificial agents which are capable of solving complex human--related tasks is one of the main challenges of Machine Learning.
When these tasks involve real--time interaction with an environment, they are usually modeled and solved with Reinforcement Learning (RL).
In Reinforcement Learning, one ore more agents interact with a stochastic environment through a discrete sequence of observations and actions.
To guide the agent towards the goal, these agents are also provided with feedback for their decisions in the form of a reward signal; the objective of an RL agent is then to maximize its cumulative reward.
Because of the generality of these concepts, Reinforcement Learning naturally applies to a high number of human--relevant tasks and to tasks involving physical interaction.

Recently, Reinforcement Learning has been leveraging the generalization power of Deep Learning, giving rise to Deep Reinforcement Learning (Deep RL) \cite{dqn}\cite{c51}\cite{doubledqn}\cite{trpo}\cite{ppo}\cite{td3}\cite{ddpg}\cite{sac}\cite{sac2}.
However, even though Deep Reinforcement Learning algorithms managed to significantly improve the state--of--the--art for a high variety of tasks, they still struggle on \emph{hard--exploration} environments \cite{goexplore}.
The challenge in solving these tasks is often not only due to the intrinsic obstacles they pose, but also to the difficulty in specifying an appropriate reward signal.
More specifically, it is often impractical or impossible to design a reward signal that can guide the RL agent through the relevant milestones needed to accomplish the goal.
Unfortunately, many challenging human--related problems are also hard--exploration problems.
This is because it is often only possible or practical to specify abstract goals for these tasks, resulting in reward functions that do not provide any clear guidance on \emph{how} the agent should achieve its goal \cite{goexplore}.
Moreover, the reward signal can even be deceiving: this happens especially when optimizing performance in the short term results in agents that are less likely to achieve the main objective of the task.

Another issue with standard Deep Reinforcement Learning algorithms is that they focus on generating single solutions to specific tasks \cite{poet}\cite{hideseek}.
For this reason, they are not directly capable of producing agents with multiple skills, or sets of diverse agents that have similar performances on a single task.

To tackle these problems, a variety of methods have been developed and studied. A first example is Intrinsic Motivation, which is based on inducing curiosity in the agent with an additional reward, unrelated to the original task.
This \emph{intrinsic} reward is designed in such a way that the agent is motivated to discover novel regions of the environment state space.

The usage of co--evolution--based algorithms has also proven to be very effective in generating and solving hard--exploration environments.
One example is the Paired Open--Ended Trailblazer \cite{poet}\cite{epoet} algorithm, which continually evolves variants of an environment, along with agents that solve them using Evolution Strategies \cite{esrl}.
POET often produces environments which are not solvable by external agents that are only trained on that environment \cite{poet}, showing that there exist tasks that are not solvable by direct optimization, and instead require that the agents go through an appropriate \emph{curriculum}.

Another notable approach is using competitiveness in multi--agent settings to produce emergent behaviour.
Even if this method usually relies on standard Deep Reinforcement Learning algorithms, the competitive interplay between multiple agents can spark complex behaviour and skills not attainable without it.
This happens when one or more agents pose adequate challenges to their opponents, resulting in a new challenge for the original opponents.
If the environment allows it, this exchange can go on indefinitely, generating more and more advanced skills in the process.
This approach of exploiting multi--agent interplay has been successfully applied to various complex games \cite{maiquake}\cite{gammon} and simple physically grounded environments \cite{sims1994a}\cite{maiquake}\cite{bansal2017}\cite{liu2019emergent}\cite{hideseek}.
\citeauthor{hideseek}, in particular, focused on the human--relevance of the skills acquired by the agents, and on the physical grounding of the simple simulated world in which they were trained.
This work also focuses on these aspects, with the double goal of: (1) developing a computationally efficient and easily extensible multi--agent, competitive environment; and (2) experimenting with variants of this environment to produce interesting emergent agent behaviours.
The first objective was achieved by developing an extensible and efficient framework, based on the Box2D physics simulation library, that allows to produce modular environments based on realistic physics.
This framework was then used to develop concrete multi--agent environments based on common survival video game semantics.
This included different levels of competitiveness, environment complexity and the subdivision of agents in opposing teams.
The second objective was partially achieved by using these concrete variants in several experiments, using standard Deep Reinforcement Learning techniques.
These experiments resulted in some simple emergent strategies, also giving concrete directions for the improvement of these environment variants in future work.

\section{Related Work}

There are several algorithms that significantly enhance the exploration capabilities of standard Deep RL algorithms.
\emph{Intrinsic Motivation} approaches introduce an additional reward signal, called \emph{intrinsic}, aimed at explicitly encouraging the agent to explore the environment.
In this context, the original reward signal is also called \emph{extrinsic}.
The intrinsic reward is given by an \emph{intrinsic reward function} $R_i$, which is in general non--stationary, in contrast with the extrinsic reward function.
Most of the Intrinsic Motivation methods shape the intrinsic reward so that it is higher in novel states, producing ``curious'' agents that seek unseen states.
One way to do this is with count--based approaches \cite{bellemare16}\cite{cbimneural}, in which the intrinsic reward is inversely proportional to the number of times a state has been visited.
This is straightforward in environments with a finite number of states, but requires additional care when the state space is infinite.
A completely different approach is to define the intrinsic reward as the prediction error for a problem related to the agent's transition \cite{peim1}\cite{peim2}\cite{peim3}\cite{peim4}\cite{rndleak}\cite{peim6}\cite{rnd}.
For example, predicting forward or inverse dynamics of the environment, or even trivial problems like predicting a constant zero function, can in practice obtain good results.
One problem of prediction--error--based Intrinsic Motivation is that the prediction error itself may be caused by several reasons, not all relating to the agent's exploration performance \cite{rnd}.
\emph{Random Network Distillation} (RND) \cite{rnd}\cite{rndleak} is a state--of--the--art Intrinsic Motivation approach that uses the prediction error on a random task, represented by a random fixed Neural Network, to compute the intrinsic reward.
While Intrinsic Motivation approaches obtain significantly better results than standard Reinforcement Learning algorithms on most hard--exploration tasks, they still leave some issues unsolved, such as \emph{detachment} and \emph{derailment} \cite{goexplore}.
Go--Explore \cite{goexplore} is an algorithm designed to solve these issues by keeping an explicit archive of interesting states and trajectories from where exploration can be resumed.

A different way to overcome hard--exploration problems with Deep RL algorithms is to provide the agent with a \emph{curriculum} of intermediate tasks, a technique called \emph{curriculum learning}.
A \emph{curriculum} in this context is a series of increasingly difficult tasks such that an agent that has solved one of them can be transferred to the next and solve it by directly maximizing its return.
Even though a curriculum can be implemented manually, this approach does not scale well with environment complexity. 
Automatic generation of appropriate curricula is an active area of Reinforcement Learning research.

One possibility is leveraging multi--agent environments.
The idea behind this approach is that, when one of the agent improves its policy and beats the other agents, they are in turn pressured to improve their own policies.
As long as these improvements don't lead to situations in which an agent is too difficult for the others to beat, this can establish an open--ended stream of increasingly difficult challenges for all agents.
This also resembles how life evolved on our planet, in that competition and co--evolution has played a fundamental role in the capabilities and diversity of living organisms.
This approach was successfully applied to complex abstract games and complex multi--player video games in \citeauthor{macomm} and \citeauthor{gammon}.
\citeauthor{sims1994a},  \citeauthor{maiquake}, \citeauthor{bansal2017}, \citeauthor{liu2019emergent} and \citeauthor{hideseek} also leveraged this approach to produce emergent behaviour using multi--agent interplay in simple physically grounded environments.

Another notable approach to automatic curriculum learning is the co--evolution of agents and tasks, as in the Paired Open--Ended Trailblazer (POET) algorithm \cite{poet}\cite{epoet}.
POET takes inspiration from minimal criterion evolution \cite{mcevo} and other Open–Ended Evolution methods \cite{oee}.
\citeauthor{pinsky} introduced a variation of POET that evolves video--game levels together with agents that play them.

\section{A Competitive Survival Environment}

The first part of this work consisted in developing a multi--agent, survival--based Reinforcement Learning environment.
The following aspects were prioritized in its design:

\begin{itemize}
    \item physical grounding of the environment dynamics, in particular classical mechanics;
    \item the possibility of changing the level of semantic complexity and competitiveness between the agents or teams;
    \item computational efficiency, such that the environment would not be a bottleneck in the experimental setup;
    \item extensibility of the implementation, so that further semantics may be implemented without modifying existing code;
    \item modularity of the implementation, so that different mechanics could be easily ``turned off'' when needed, or combined in different ways.
\end{itemize}

This section specifies the dynamics and semantics of the developed environment, while appendix \ref{s:impl} discusses their implementation, along with the framework developed to make them extensible and modular.

The environment is set in a planar world where bodies follow the laws of classical mechanics.
The relevant region of the plane is a \emph{room} enclosed by four walls.
The environment can contain one or more agents, all acting simultaneously.
Agents can interact with other entities in the environment in several ways, including taking damage from various sources.
The semantics of these interactions are detailed in sections \ref{ss:agents}, \ref{ss:itemsobjs} and \ref{ss:safezone}, while section \ref{ss:envspaces} defines the action and observation spaces for each agent.
Finally, section \ref{ss:rewdone} discusses the reward schemes and episode termination conditions.
Note that a multitude of variants of this environment are possible; a small fraction of these possibilities are depicted in figures \ref{fig:screen1}, \ref{fig:screen2} and \ref{fig:screen3}.

\newcommand\screenw{0.45}

\begin{figure}[t]
\centering
\includegraphics[width=\screenw\textwidth]{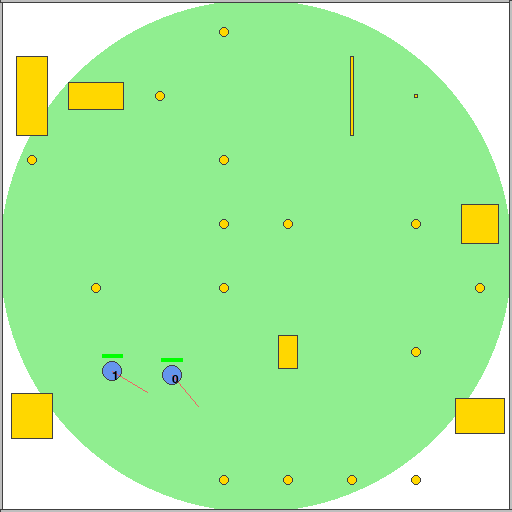}
\caption{Screenshot of an environment variant with heals and boxes and 2 agents. The segment originating from the agents is the melee, the green bar on top of them is their health, while the number appearing on their side is their agent ID.}
\label{fig:screen1}
\end{figure}

\begin{figure}[t]
\centering
\includegraphics[width=\screenw\textwidth]{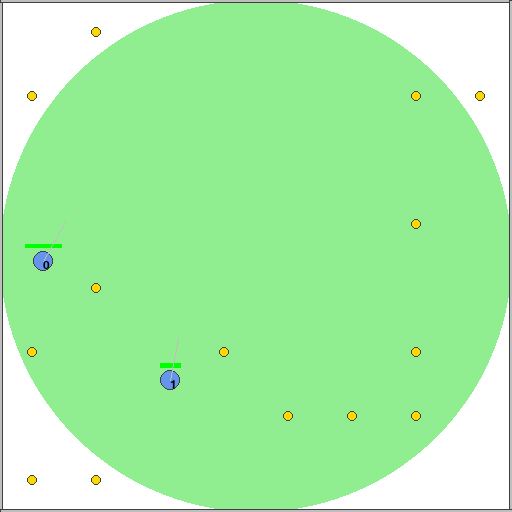}
\caption{Screenshot of an environment variant with only heals and 2 agents.}
\label{fig:screen2}
\end{figure}

\begin{figure}[t]
\centering
\includegraphics[width=\screenw\textwidth]{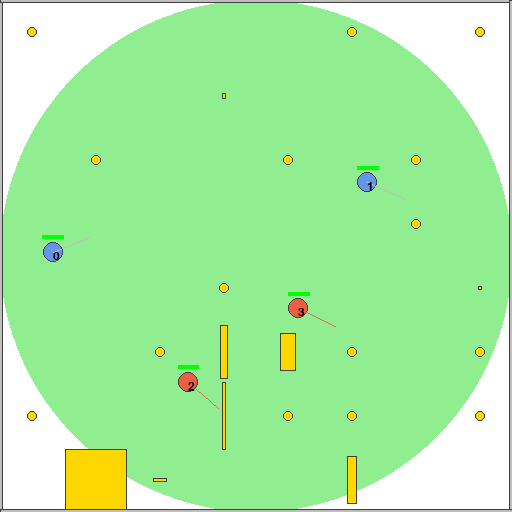}
\caption{Screenshot of an environment variant with heals, boxes, and agents divided into teams.}
\label{fig:screen3}
\end{figure}

\subsection{Agents} \label{ss:agents}

Agents are modeled as circles with a fixed radius. They can apply forces to their bodies in the directions parallel and orthogonal to their orientation, along with torque along the axis orthogonal to the room plane. All agents have an integer health; when their health becomes 0 or less, they despawn and are considered dead. Agents can attack other bodies, including other agents, by using a \emph{melee}, a close--ranged weapon. This is modeled as a segment of fixed length and parallel to the direction the agent is facing. An attack succeeds only if the target touches the melee.

Agents can also carry items in a small inventory with a fixed number of slots. The item in the last slot can be either used, dropped on the ground, or given to another agent that is sufficiently close. Moreover, when an agent dies, the contents of their inventory are dropped on the ground, randomly scattered in a small circular region centered on the last position of the agent. Items are described more in detail in section \ref{ss:itemsobjs}.

Depending on the environment variant, agents can also be grouped into teams. In this case, agents can only attack other agents outside of their team and they can only give items to agents inside their team. Moreover, agents in the same team completely share their rewards.

\subsection{Items and Objects} \label{ss:itemsobjs}

Items are modeled as small, intangible circular bodies. Agents can pick them up in their inventory by touching the item body, which then despawns. Each item has a specific behaviour when it is used. This environment includes two kinds of items: \emph{heals} and \emph{object items}. Heals are items that simply increase the health of their user. Object items are items that spawn other bodies, such as boxes, at the location of the user. These objects can be broken to drop an item that respawns them. In this environment, objects are rectangular boxes; the dimensions of each box are determined randomly at the start of each episode.

\subsection{The Safe Zone} \label{ss:safezone}

The \emph{safe zone} is a popular mechanic in competitive ``Battle--Royale'' video games. It is a moving region of the map that shrinks as the game advances, hurting all players that are outside of it with a small but constant damage. Thus, not staying inside the safe zone is eventually lethal. Moreover, the safe zone usually shrinks to a void region near the end of the game, so that players cannot survive indefinitely. The safe zone does not by itself introduce competitiveness in the environment; however, if paired with an appropriate reward scheme, it indirectly forces agents to fight for their spot in the safe zone to stay alive longer.

In this environment, the safe zone is circular and alternates between a stationary phase and a shrinking--moving phase. The positions of the center for each phase are extracted randomly, such that the zone is always fully inside of the room. The radius of the zone decreases linearly during each shrinking--moving phase, until it becomes null after the last phase.

\subsection{Observation and Action Spaces} \label{ss:envspaces}

The observation $x$ available to each agent is in the form
\[ x = (x_{\text{self}}, x_{\text{zone}}, x_{\text{entities}}) \]
All the components are described in more detail below. All the scalar values of the observation are real numbers, so that the observation space is continuous.

$x_{\text{self}}$ contains the information about the observing agent, and is in the form
\[ x_{\text{self}} = (i, h, x, y, \theta, v_x, v_y, \omega) \]
where:
\begin{itemize}
    \item $i$ is the ID of the agent;
    \item $h$ is its current health;
    \item $p = (x, y)$ is its absolute position;
    \item $\theta$ is its orientation;
    \item $v = (v_x, v_y)$ is its velocity;
    \item $\omega$ is its angular velocity.
\end{itemize}
Additionally, if the agents are divided into teams, $x_{\text{self}}$ also contains the ID of the team after the agent ID.

$x_{\text{zone}}$ observes the current center and radius of the safe zone, along with the center and radius of the next phase of the safe zone. More precisely, it is in the form
\[ x_{\text{zone}} = (c_x, c_y, r, c'_x, c'_y, r') \]
where $(c_x,c_y)$ is the center of the current safe zone, $r$ its radius, and $(c'_x,c'_y)$ and $r'$ are the center and radius of the safe zone in its next stationary phase.

The remaining component of the observation, $x_{\text{entities}}$, contains all the information about the various entities that are in the environment, and is in the form
\[ x_{\text{entities}} = (x_{\text{other}},x_{\text{heal}},x_{\text{box}},x_{\text{boxitem}},x_{\text{healslot}},x_{\text{boxslot}}) \]
All the components $x_{\text{entities}_i}$ of $x_{\text{entities}}$ have a similar structure but different sizes, and each correspond to a different type of entity in the environment. Each $x_{\text{entities}_i}$ is in the form
\[ x_{\text{entities}_i} = (x_{\text{allents}_i}, x_{\text{mask}_i}) \]
$x_{\text{allents}_i}$ is a matrix which contains, for each entitity of that type, the actual information about it, stored as rows. The $x_{\text{mask}_i}$ component, instead, is a single vector that contains binary values. Each of these values corresponds to a row of $x_{\text{allents}_i}$ and tells whether the observing agent can see that entity with its camera or not. The dimension and contents of each row of $x_{\text{allents}_i}$ depend on the entity type as follows:
\begin{itemize}
    \item For $x_{\text{other}}$, the entities are the other agents; each row is thus in the same form as $x_{\text{self}}$.
    \item For $x_{\text{heal}}$, entities are the heal items, and each row contains the position $(x,y)$ of the item.
    \item For $x_{\text{box}}$, the entities are the box objects. Each entity row contains: the 2D positions of each of its 4 vertices, as if the box was centered at the world origin; the translation vector to the actual center of the box in the world; the actual orientation of the box in the world.
    \item For $x_{\text{boxitem}}$, the entities are the items corresponding to broken boxes. Each row contains: the 2D positions of each its 4 vertices, as if the box was centered at the world origin; and the position of the box item in the world.
    \item $x_{\text{healslot}}$ always contains a single entity, which represent the last slot of the agent's inventory, if it contains a heal item. Otherwise, its only row is masked out in the corresponding $x_{\text{mask}_{\text{healslot}}}$.
    \item Similarly, $x_{\text{boxslot}}$ also contains a single entity, which represent the last slot of the agent's inventory, if it contains a box item. Otherwise, its only row is masked out in the corresponding $x_{\text{mask}_{\text{boxslot}}}$.
\end{itemize}

The action space of the environment is discrete, and each action $a$ is in the form
\[ a = (a_x, a_y, a_\theta, a_{\text{atk}}, a_{\text{use}}, a_{\text{give}}) \]
where:
\begin{itemize}
    \item $a_x$, $a_y$ and $a_\theta$ all take values in $\{-1, 0, 1\}$, and give the direction of the linear and angular force that the agent can apply to itself; their value is multiplied by a constant parameter that controls the magnitude of the resulting forces
    \item $a_{\text{atk}}$ is a binary action that controls whether the agent is trying to attack with its melee at the current time step;
    \item $a_{\text{use}}$ is a binary action that, when active, makes the agent try to use the last item it picked up in its inventory;
    \item $a_{\text{give}}$ is a binary action, and controls whether the agent is trying to give the last item it picked in its inventory to the nearest suitable agent or teammate. 
\end{itemize}

\subsection{Rewards and Episode Termination} \label{ss:rewdone}

The reward function $R$ of a single agent or team is deterministic, depends only on the states $s$ and $s'$, and is defined as

\[ R_i(s, s') = I_{\text{alive}}(s, s') r_{\text{alive}} + (1 - I_{\text{alive}}(s, s')) r_{\text{dead}} + n_{\text{kills}}(s, s') r_{\text{kill}} + I_{\text{death}}(s, s') r_{\text{death}} \]
where:
\begin{itemize}
    \item $I_{\text{alive}}(s, s')$ is 1 when the agent is alive, and 0 otherwise;
    \item $I_{\text{death}}(s, s')$ is 1 if the agent died in the transition from $s$ to $s'$, and 0 otherwise;
    \item $n_{\text{kills}}(s, s')$ is the number of enemy agents the agent has successfully killed in the transition from $s$ to $s'$;
    \item $r_{\text{alive}}$ is a parameter controlling the reward obtained by live agents at each step;
    \item $r_{\text{dead}}$ is a parameter controlling the reward obtained by dead agents at each time step;
    \item $r_{\text{kill}}$ is the reward obtained by each successful kill;
    \item $r_{\text{death}}$ is the reward obtained at the time step at which the agent dies.
\end{itemize}
Each environment variant can specify different numbers for the four parameters.

There are two possible termination conditions for an episode, depending on the environment variant. The first is to consider an episode terminated when all agents or teams are dead. The second is to end an episode when there is only one or less agents or teams still alive. Along with the four reward parameters described above, this choice determines the level of competitiveness of the environment. For example, consider an environment in which only $r_{\text{alive}} = 1$, while all other parameters are 0, and the episode terminates when all agents are dead. In this environment, there is no significant advantage is adopting an aggressive strategy and attacking other agents, as long as the safe zone is big enough. The best strategy is simply to survive as long as possible. On the other hand, if $r_{\text{kill}}$ is positive and the episode ends when only one agent or team is alive, there is considerable advantage in killing opponents to collect reward. Note that depending on the precise value of $r_{\text{kill}}$, it may also be convenient to survive as long possible before engaging opponents to collect the kill reward.

\section{Experiments and Results} \label{s:setup}

This section details the experimental setup and the results of all the reported experiments. The following paragraphs discuss some aspects common to all experiments.
Section \ref{ss:policyopt} describes the policy optimization approach, while subsequent sections contain experiment results.

The three main aspects that differentiate the environment variants used for these experiments are:

\begin{itemize}
    \item whether the agents are grouped into teams or not;
    \item competitiveness of the reward scheme and episode ending conditions;
    \item complexity of the environment.
\end{itemize}

Two teaming modes were used here: free--for--all, and division of the agents into 2 opposite teams.
The competitiveness varied among:

\begin{itemize}
    \item \emph{low} competitiveness: $r_{\text{alive}} = 1$, all other reward parameters set to 0, and episodes ending only when all agents were dead;
    \item \emph{medium} competitiveness: $r_{\text{alive}} = 1$, $r_{\text{dead}} = -1$, the other two reward components set to 0, and episodes ending when only agent was still alive;
    \item \emph{high} competitiveness: $r_{\text{alive}} = 1$, $r_{\text{dead}} = -1$, $r_{\text{kill}} = 100$, $r_{\text{death}} = -100$.
\end{itemize}

The complexity of the environment mainly involved the presence or absence of randomly shaped boxes.

Besides the rewards obtained by every agent or team, the following variables were also recorded for each episode of experience during training:
\begin{itemize}
    \item the number of heal items consumed;
    \item the number boxes placed by consuming their item;
    \item the number of kills caused by every agent or team.
\end{itemize}

Performance tests were performed on all variants described below, with the results reported in table \ref{tab:expperf}.


\subsection{Policy Optimization} \label{ss:policyopt}

\begin{table}[b]
    \centering
    \caption{Policy optimization hyperparameters.}
    \label{tab:hyperparams}
    \begin{tabular}{|c|c|}
        \hline
        \textbf{Hyperparameter} & \textbf{Value} \\
        \hline
        Rollout buffer size $T$ & $2^{14} = 16384$ \\
        Minibatch size $B$ & $2^9 = 512$ \\
        PPO Clip parameter $\epsilon$ & 0.2 \\
        Discount factor $\gamma$ & 0.99 \\
        GAE parameter $\lambda$ & 0.95 \\
        Target KL divergence $\delta$ & 0.01 \\
        \hline
    \end{tabular}
\end{table}

The policies of the agents are represented with Artificial Neural Network models.
Each agent has two networks associated to it: the defines a stochastic policy by outputting action distributions; the second is used to approximate the value function.
The architectures of the agent networks are based on fully--connected layers and masked self--attention to account for the varying number of entities observed by each agent.
These models were optimized using the Proximal Policy Optimization algorithm \cite{ppo} and Generalized Advantage Estimation \cite{gae}.
A version of the StableBaselines 3 implementation of PPO \cite{sb3} was used, modified to optimize multiple agents at once. This implementation features early stopping during each step of policy optimization, based on an estimate of the Kullback--Leibler divergence between the old and new policies.
Early experiments in this environment showed that this was a key aspect of the policy optimization algorithm, and adopting a fixed number of SGD steps instead of early stopping produced either destructive updates to the policies, or much slower learning.
Table \ref{tab:hyperparams} reports the values of the hyperparameters used for policy optimization.

As is common in multi--agent Reinforcement Learning literature, the parameters of the agent networks were shared between all agents \cite{hideseek}.
Contrary to the approaches in \citeauthor{hideseek}, \citeauthor{omniscent1}, \citeauthor{omniscent2} and \citeauthor{omniscent3}, in which agents are trained on ``omniscent'' observations, here non--visible entities were masked also during training.
This choice was made after early experiments revealed that agents trained with omniscent observations were slightly less prone to explore their surroundings in search for items, e.g. heals.

The architectures of the policy and value networks used for the agents are very similar, with the only notable difference being the last few layers.
Remember that the agent observations $x$ are in the form
\[ x = (x_{\text{self}}, x_{\text{zone}}, x_{\text{entities}}) \]
And each $x_{\text{entities}_i}$ is in the form
\[ x_{\text{entities}_i} = (x_{\text{allents}_i}, x_{\text{mask}_i}) \] 
The inputs $x_{\text{self}}$ and $x_{\text{zone}}$ are concatenated and fed to a fully--connected layer, obtaining the non--entity embedding $z$.
For each of the entity types, the corresponding rows of $x_{\text{allents}_i}$ are all passed through the same fully--connected layer, with output dimension $N_e$ shared between all entity types.
In this way, all the entity embeddings can be concatenated into a single sequence of embedded entities $E$.
Then, the masks $x_{\text{mask}_i}$ for all entity types are concatenated to obtain a mask that refers to entity rows in $E$.
$E$ and this mask are then passed through a Self--Attention layer, which is an Attention layer where queries, keys and values all come from the same sequence.
The output sequence of this Attention layer is then passed through an average pooling to obtain a vector of fixed length $z_{\text{ent}}$, which does not depend on the number of entities.
$z$ and $z_{\text{ent}}$ are then concatenated and fed through two fully--connected layers.
The policy network then has $N_a$ fully--connected heads, one for each action, each with output dimension equal to the number of possible values for that action.
A softmax activation is then used to output an action distribution for each action component. 
The value function network, instead, simply has one fully--connected head with a single output and no activation function. All fully--connected hidden layers use ReLU activations.

\subsection{Free--for--all}


\newcommand\plotw{0.8}

\begin{figure}[t]
\centering
\includegraphics[width=\plotw\textwidth]{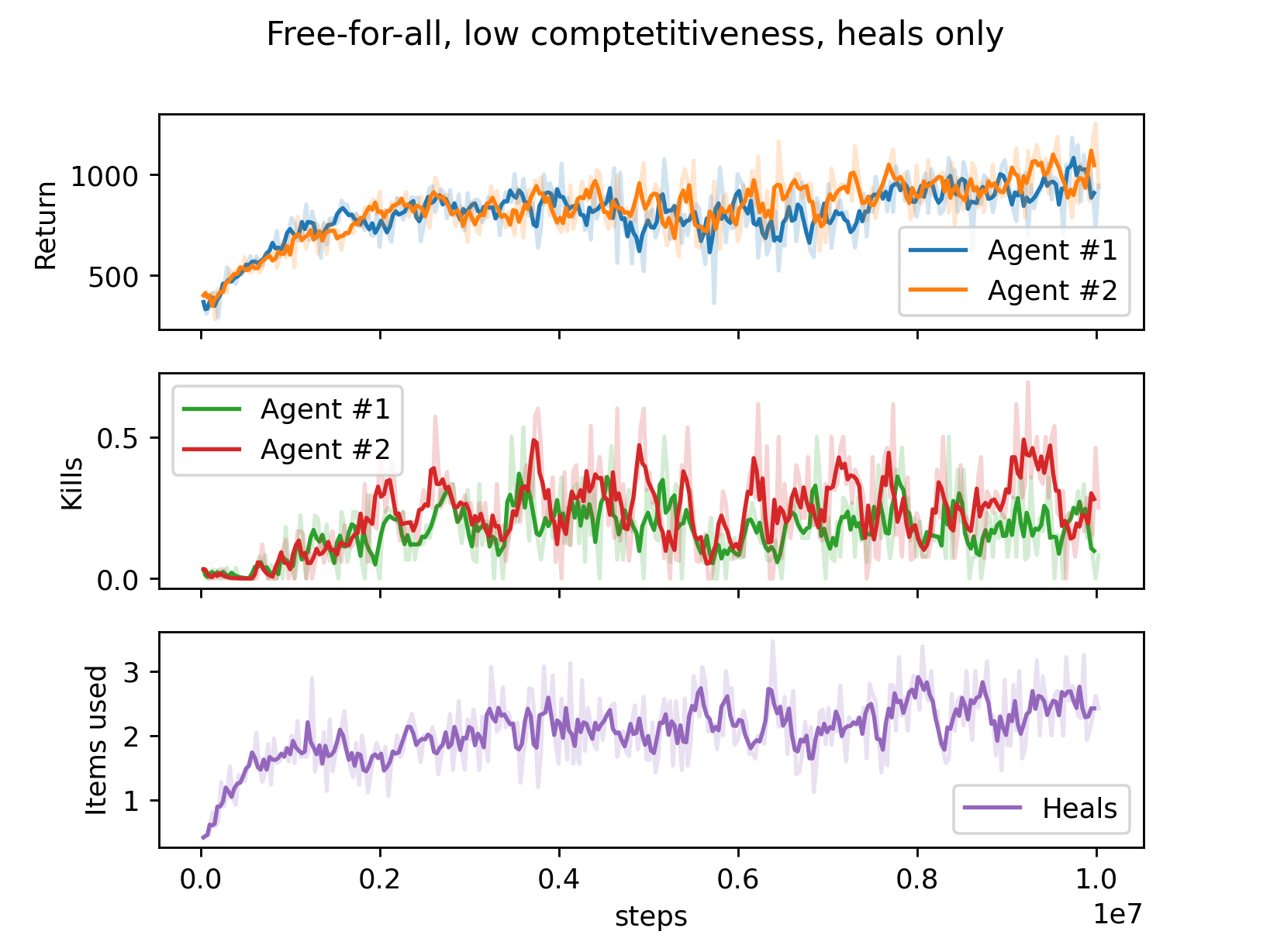}
\caption{Results for the free--for--all variant 1. The number of kills, while still lower than 1, is still significantly higher than other free--for--all variants. Its trend is to increase in the early stages of training and stabilize to a very noisy but seemingly constant value.}
\label{fig:ffa1}
\end{figure}

\begin{figure}[t]
\centering
\includegraphics[width=\plotw\textwidth]{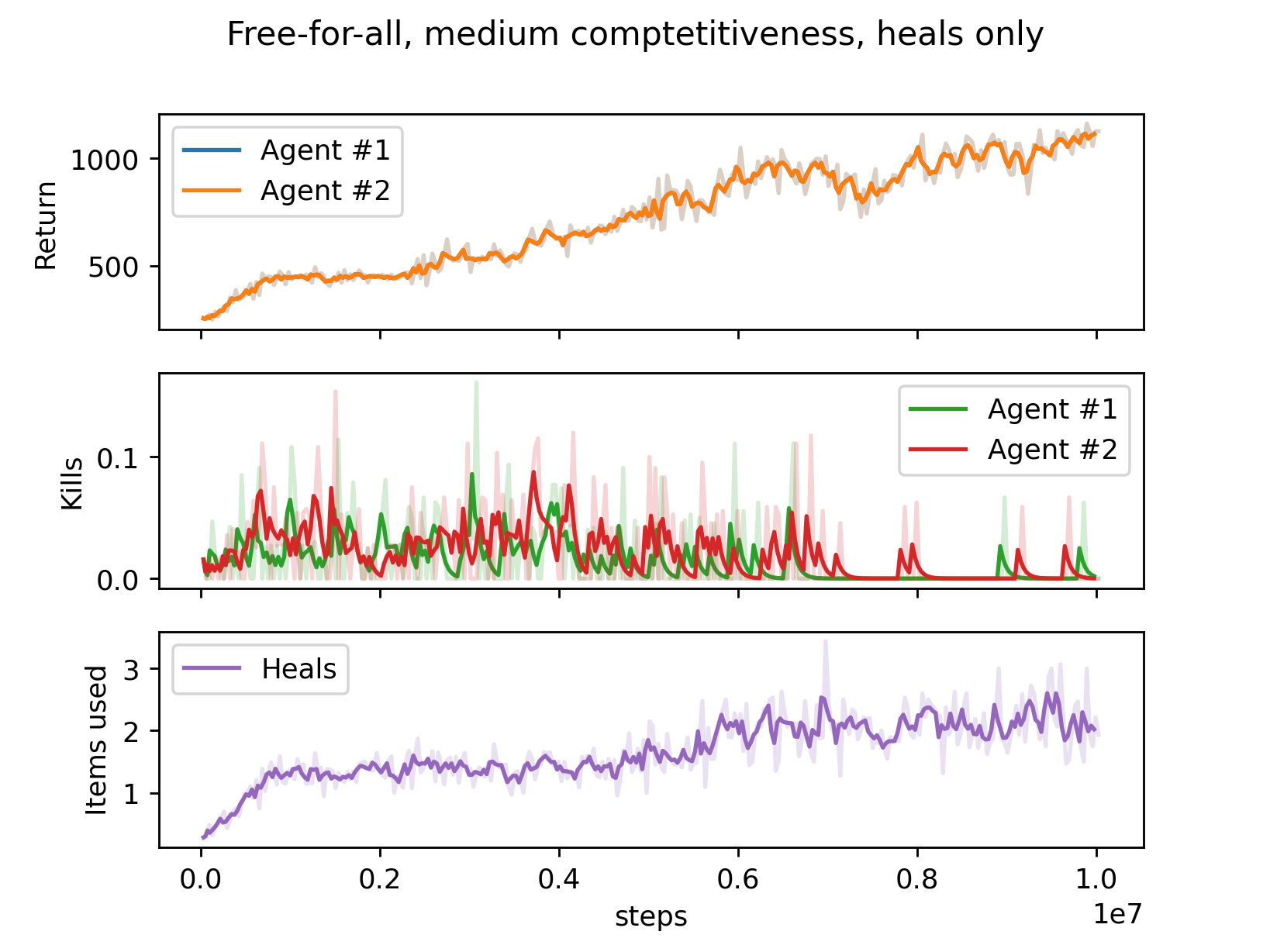}
\caption{Results for the free--for--all variant 2. Here, as agents learn to avoid hurting their opponent, the number of kills almost vanishes in the last stages of training. Note that the reward scheme results in the returns only differing by 1, which so that the two return lines seem to coincide in the plot.}
\label{fig:ffa2}
\end{figure}

\begin{figure}[t]
\centering
\includegraphics[width=\plotw\textwidth]{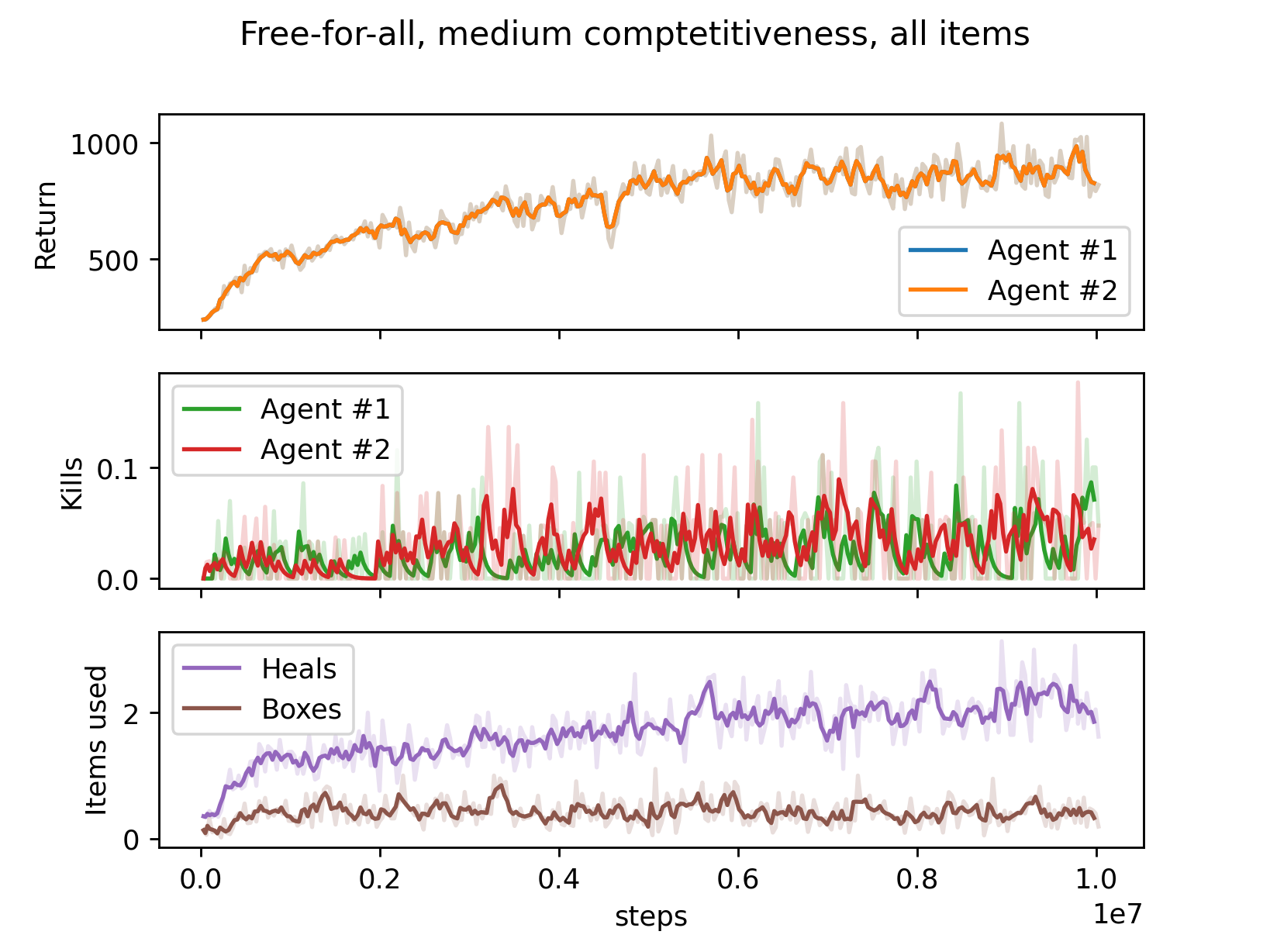}
\caption{Results for the free--for--all variant 3. The number of kills increases very slowly, as it does its variance; however, its value still remains very close to 0. The number of boxes placed remains close to 0 for the whole training process. Note that the reward scheme results in the returns only differing by 1, which so that the two return lines seem to coincide in the plot.}
\label{fig:ffa3}
\end{figure}

\begin{figure}[t]
\centering
\includegraphics[width=\plotw\textwidth]{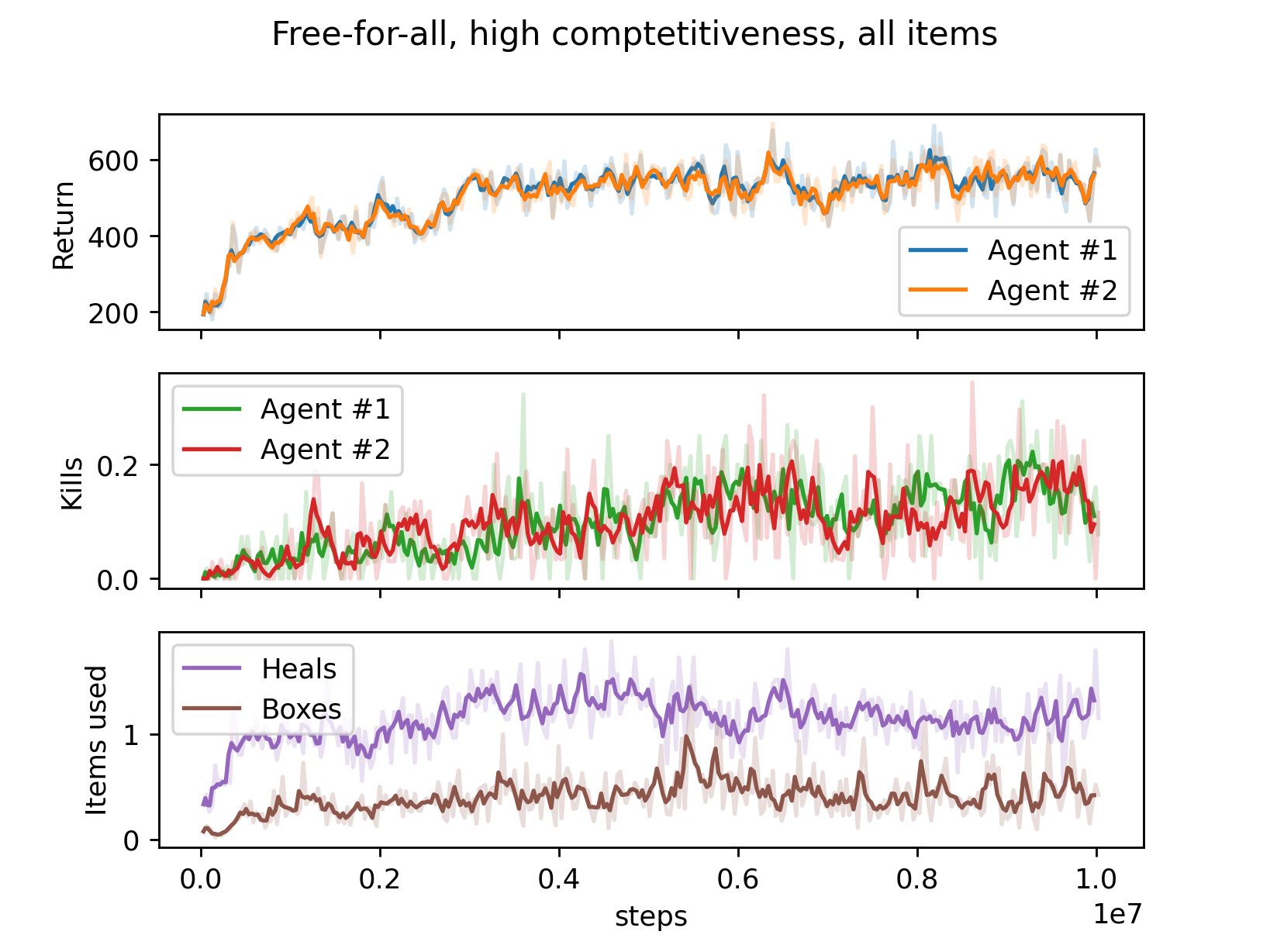}
\caption{Results for the free--for--all variant 4. The number of kills slowly increases, still remaining close to 0, but with a more consistent trend than variant 2. The number of boxes placed also remains close to 0, and the number of heals is significantly lower than other variants.}
\label{fig:ffa4}
\end{figure}

The first group of experiments performed involved variants of the environment in which all agents are against all other agents. In all variants, the number of agents was 2. The variants are as follows:
\begin{enumerate}
    \item free-for--all with low competitiveness and only heal items;
    \item free-for--all with medium competitiveness and only heal items;
    \item free-for--all with medium competitiveness, heal items and boxes;
    \item free-for--all with high competitiveness, heal items and boxes;
\end{enumerate}

All variants were allocated 10 million environment steps for training. The figures \ref{fig:ffa1}, \ref{fig:ffa2}, \ref{fig:ffa3} and \ref{fig:ffa4} show the results for the 4 free--for--all variants. In all of them, the reward and the number of heals used steadily increase for the whole training process. This suggests that all the agents have learned the basic strategy of following the safe zone and collecting heals to survive longer. To confirm this, the policies learned by agents in all 4 variants were visually inspected over multiple instances of the environment, giving rise to the following observations:
\begin{enumerate}
    \item In all variants, the agents learn to follow the safe zone and collect heal items when they are in their line of sight. However, in variant 4 the agents apply this strategy much more inconsistently, often moving around outside the zone, or failing to collect heals in their line of sight. In other variants, where episodes often get to the point where the safe zone disappears, agents have also learned to move frantically to try to find heal items, so that they can survive a little bit longer.
    \item In variants 2 and 3, agents learn to avoid hurting their opponent by not attacking them when they are in range (note that if an agent is in the attack range of another, it is also always in its vision cone). Moreover, they do so in different ways: in variant 3, which also features boxes, agents only use the attack action when a box is in their attack range. Instead, in variant 2, the agents seem to ``spam'' the attack action when no agent is in range, but almost stop using it when an opponent is in range.
    \item In variant 1, where there is no clear incentive for competitiveness, agents don't seek the opponent intentionally, but almost always use the attack action if they find an opponent is in range.
    \item In variant 4, the outcome of episodes is very inconsistent, and mostly alternates between three kind of behaviour: (1) one agents kills the other in the initial phases, resulting in poor rewards; (2) episodes in which agents don't interact and die because they are eventually forced outside the safe zone, obtaining high reward; (3) episodes in which agents die because of inconsistency in following the safe zone.
    \item In all variants, agents sometimes struggle with motion control, occasionally wandering outside the safe zone for a small period of time, or missing a heal item because their speed was too high.
\end{enumerate}

\subsection{Teams}




\begin{figure}[t]
\centering
\includegraphics[width=\plotw\textwidth]{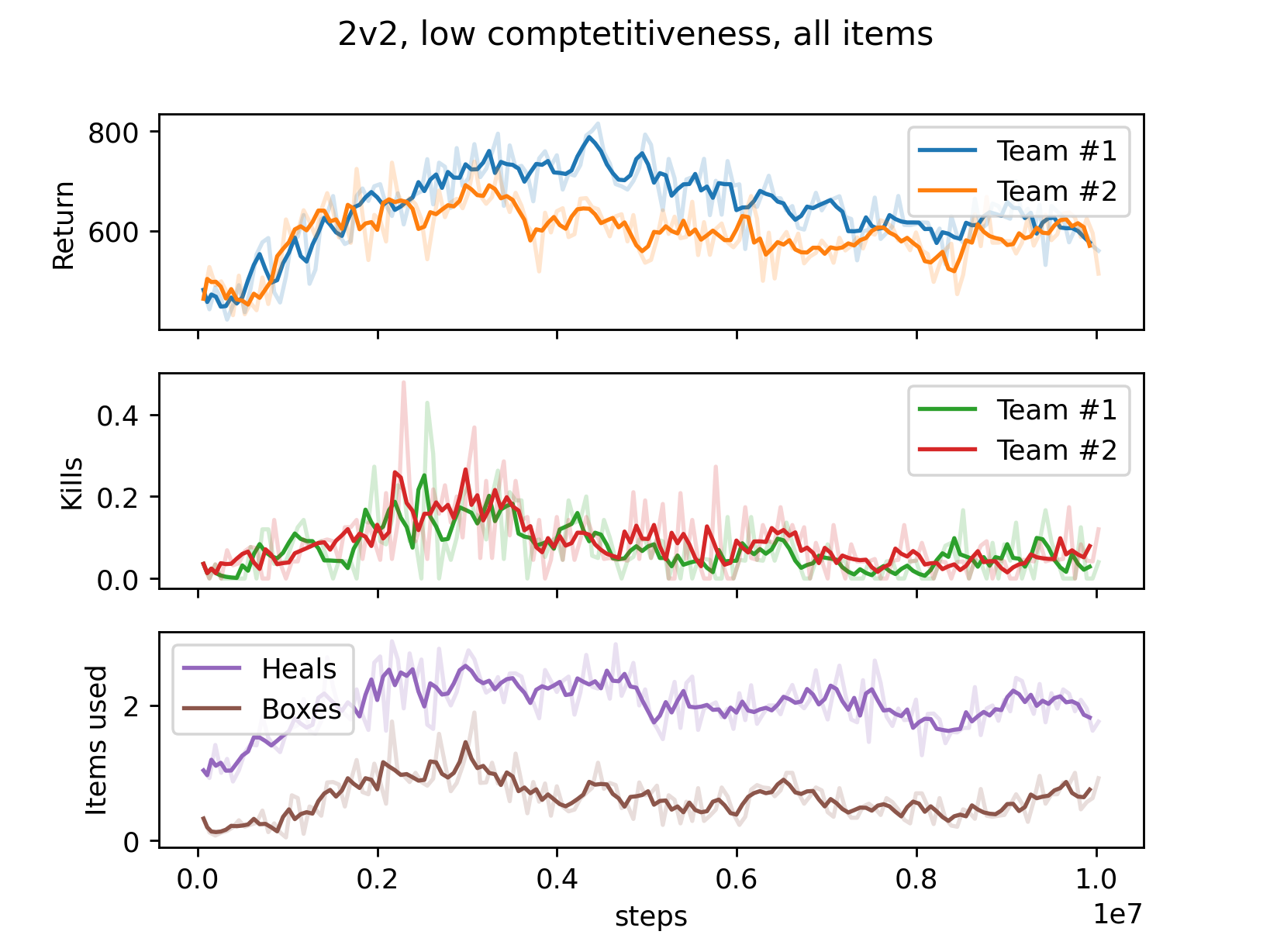}
\caption{Results of the 2 vs 2 variant 1. The number of placed boxes and kills peaks at around 3 million steps, while the peak return is observed later at 4 around million steps. The number of kills, in particular, still takes values close to 0, but significantly higher than varinats 2 and 3. The number of heals used initially increases, and the follows a very slow decreasing trend after around 4 million steps.}
\label{fig:teams1}
\end{figure}

\begin{figure}[t]
\centering
\includegraphics[width=\plotw\textwidth]{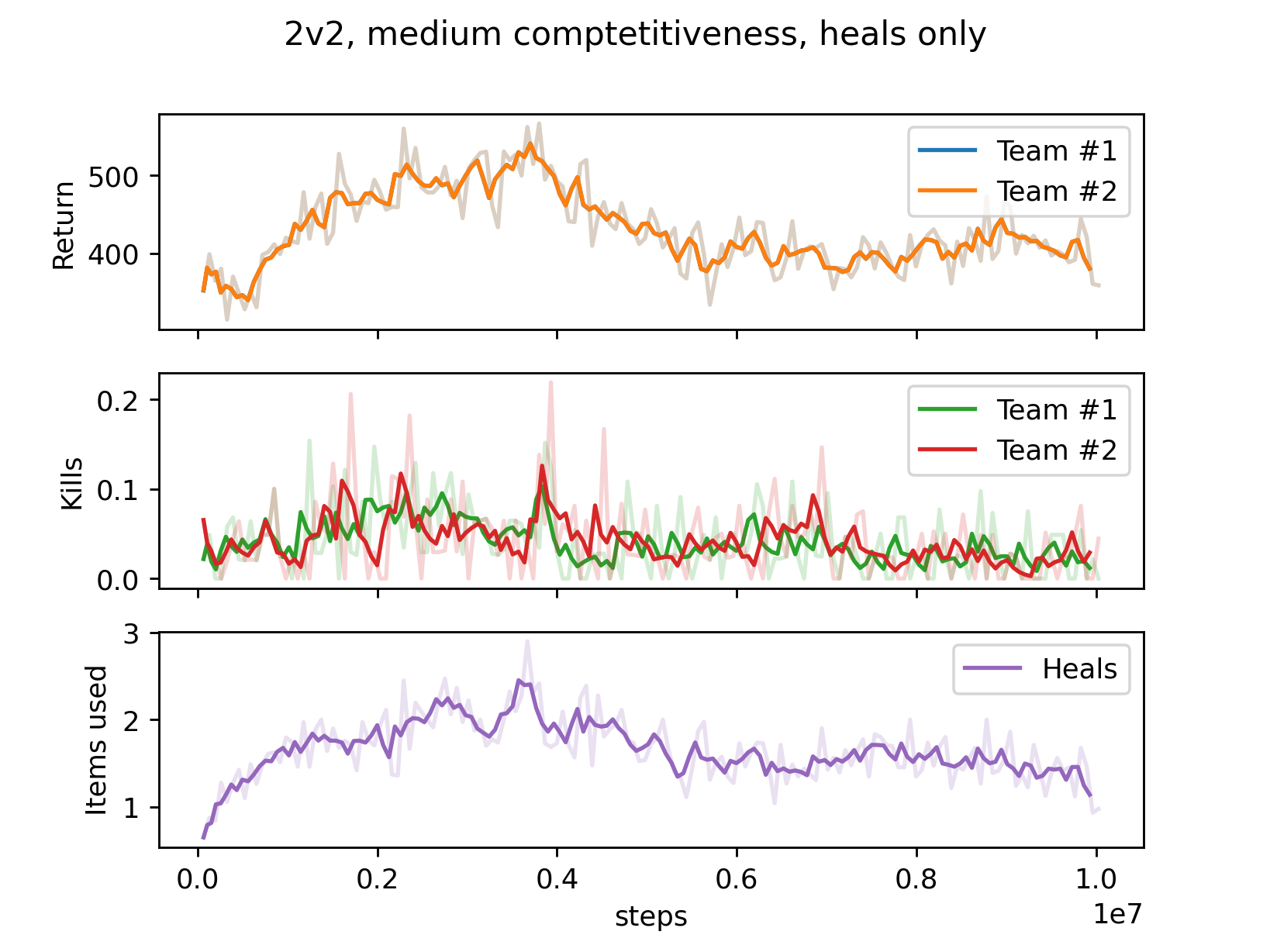}
\caption{Results of the 2 vs 2 variant 2. Again, the episode return peaks around 4 million episodes. Kills have a noisy but higher value from 2 million steps to 4 million steps, and then slowly decrease until training ends. Nonetheless, kills always take values closer to 0 than 1. Heals follow the reward trend. Note that the reward scheme results in the returns only differing by 1, which so that the two return lines seem to coincide in the plot.}
\label{fig:teams2}
\end{figure}

\begin{figure}[t]
\centering
\includegraphics[width=\plotw\textwidth]{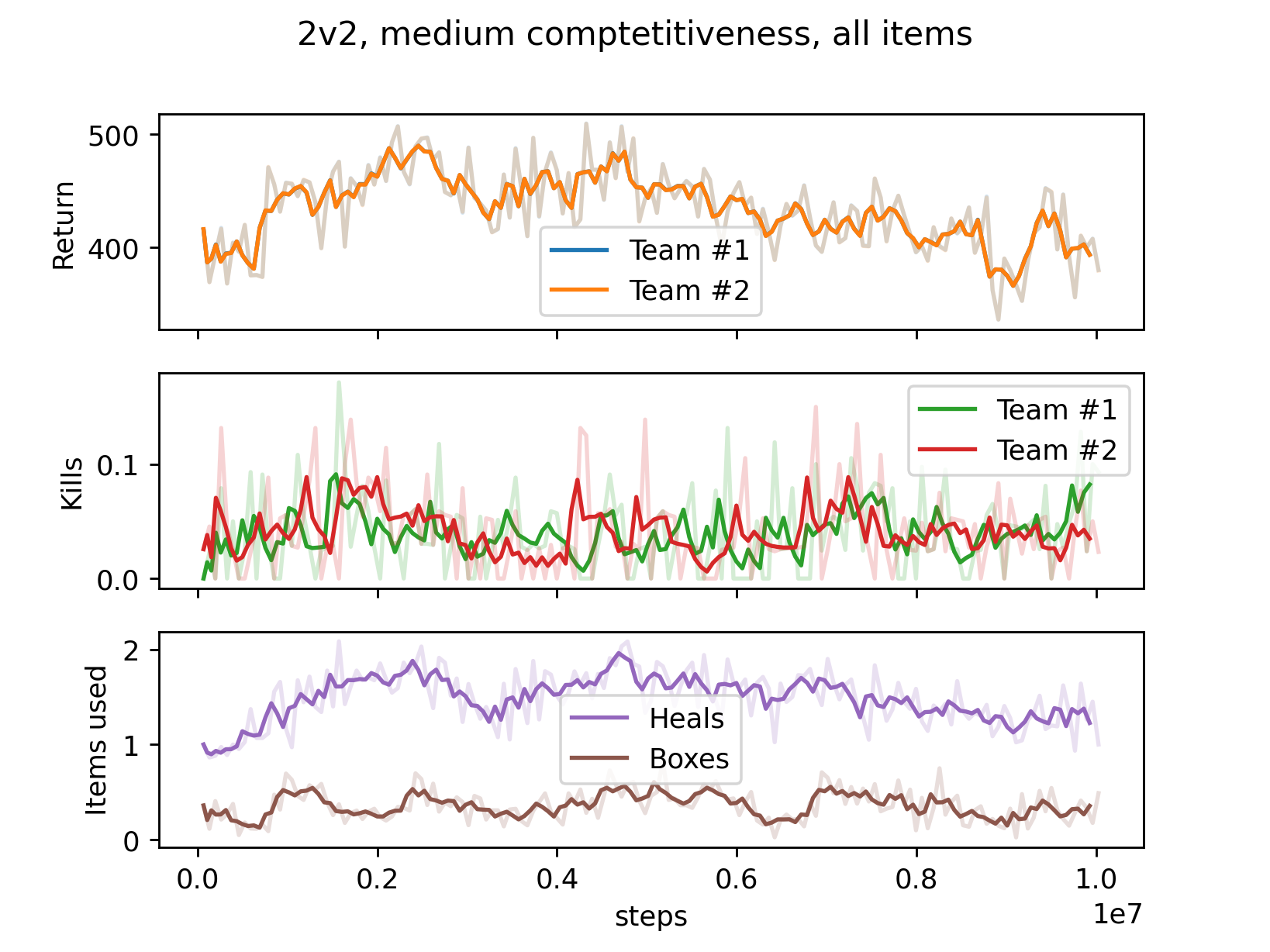}
\caption{Results of the 2 vs 2 variant 3. The trends are not as clear as the first 2 variants, but still show the return initially increasing and then degrading in later stages of training. The number of kills is noisy but still remains smaller than the first 2 variants and close to 0, not showing any clear trend. Note that the reward scheme results in the returns only differing by 1, which so that the two return lines seem to coincide in the plot.}
\label{fig:teams3}
\end{figure}

\begin{figure}[t]
\centering
\includegraphics[width=\plotw\textwidth]{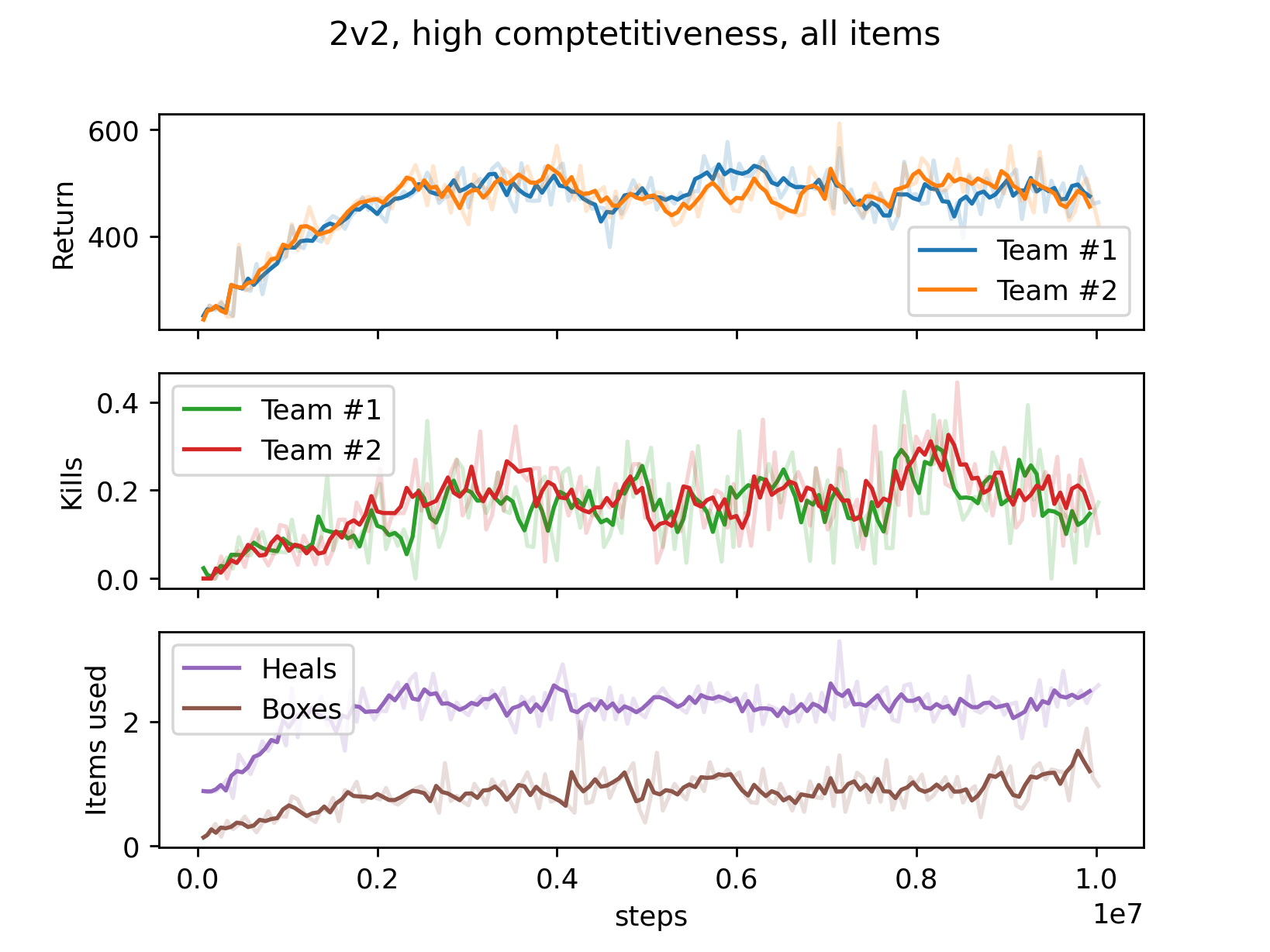}
\caption{Results of the 2 vs 2 variant 4. The kills show a a slow increasing trend throughout the training, taking values of magnitude comparable to that of variant 1. The number of boxes placed shows a very slow but clear increasing trend, while the number of heals used stabilizes at around 2 million steps.}
\label{fig:teams4}
\end{figure}

In the second group of experiments, agents were divided into two teams with shared reward and episode termination conditions. All variants feature 4 agents, so that each team was composed of 2 agents. The variants are as follows:
\begin{enumerate}
    \item 2 vs 2 with low competitiveness, heal items and boxes;
    \item 2 vs 2 with medium competitiveness, heals only;
    \item 2 vs 2 with medium competitiveness, heal items and boxes;
    \item 2 vs 2 with high competitiveness, heal items and boxes;
\end{enumerate}

As for the free--for--all variants, the training lasted until 10 million environment steps were reached. Results are shown in figures \ref{fig:teams1}, \ref{fig:teams2}, \ref{fig:teams3} and \ref{fig:teams4}. The reward trend for all these variants, besides variant 4, is an initial increase until around 4 million environment steps. Afterwards, the rewards of both teams decrease significantly. The number of heals used follows the reward trend for all four variants. In variants 1 and 4, the number of boxes placed reaches values slightly greater than 1, even though it decreases back to smaller values in variant 1. The trend of kills per episode is heavily dependent on the variants, but doesn't reach values close to 1 for any variant. For the first 3 variants, the visual inspection of the policies has been performed both with the parameters values at 10 million steps and with earlier parameter values at 4 million steps. These are the resulting observations:
\begin{itemize}
    \item At 10 million steps, in the first 3 variants agents have learned to inconsistently follow the safe zone and collect and use heal items. The inconsistency is mainly due to the agents moving around, often spreading out in different corners of the map. The agents don't intentionally seek opponents, but almost always attack them when they are in range. It is also often the case that one agent on each team dies fairly early in the episode by standing outside the safe zone, while its teammate manages to survive much longer. Note that in this case, the death of the first teammate does not directly affect the episode return for the team.
    \item At 4 million steps, the agents of the first 3 variants follow the safe zone much more consistently than at 10 million steps. Moreover, while it is not clear whether this is intentional, agents seem to find themselves more grouped towards the center of the safe zone. Moreover, it is less frequent that one of the teammates of each team dies in the first steps of the episode by standing outside the safe zone. As for the policies at 10 million steps, there is no intentional seeking for opponents, although agents almost always use the attack action on opponents (and possibly boxes) in range.
    \item In variant 4, the agents inconsistently follow the safe zone and collect heal items. As in other variants, they attack opponents and boxes when in range, but don't actively seek opponents to kill. Also in this case, the inconsistency comes from agents occasionally wandering around the map, stepping outside the safe zone.
\end{itemize}

\begin{table}[!b]
    \centering
    \caption{Computational performance tests on the environment step function. Each result reports the mean and standard deviation of the time taken by the \texttt{env.step} call, over the first 100 steps of an episode. These are usually the most expensive steps, since all agents and objects are still in the world, so the estimates reported are pessimistic. The dependence on the objects present in the world is also confirmed by the fact that configurations which feature less agents or the absence of boxes have obtain faster times. Moreover, configurations with the same number of objects show almost perfectly overlapping results. The tests were run on a Macbook Pro '15 laptop, with an Intel Mobile Core i7 ``Broadwell'' (I7-5557U) CPU.}
    \label{tab:expperf}
    \begin{tabular}{|c|c|}
        \hline
        \textbf{Variant} & \textbf{Seconds per step ($\mu \pm \sigma$)} \\
        \hline
        Free--for--all variant 1 & $7.59\cdot{10^{-4}} \pm 1.1\cdot{10^{-4}}$ \\
        Free--for--all variant 2 & $7.58\cdot{10^{-4}} \pm 1.2\cdot{10^{-4}}$ \\
        Free--for--all variant 3 & $9.86\cdot{10^{-4}} \pm 6.28\cdot{10^{-5}}$ \\
        Free--for--all variant 4 & $1.04\cdot{10^{-3}} \pm 2.97\cdot{10^{-5}}$ \\
        2 vs 2 variant 1 & $1.47\cdot{10^{-3}} \pm 1.48\cdot{10^{-4}}$ \\
        2 vs 2 variant 2 & $1.21\cdot{10^{-3}} \pm 1.30\cdot{10^{-4}}$ \\
        2 vs 2 variant 3 & $1.40\cdot{10^{-3}} \pm 7.25\cdot{10^{-5}}$ \\
        2 vs 2 variant 4 & $1.42\cdot{10^{-3}} \pm 6.08\cdot{10^{-5}}$ \\
        \hline
    \end{tabular}
\end{table}

\section{Discussion} \label{s:discussion}


In all experiments, agents learned the basics of the environment: staying inside the safe zone, and collecting and using heal items. However, there is high variability in how consistently the agents follow this strategy, depending on the variant. The first 3 free--for--all variants are the only ones in which agents reliably follow this strategy. The causes seem to depend on the variant.

For the 2 vs 2 variants in which the return decreases in the last stages of training, one possibile cause is some form of catastrophic forgetting. However, observations of the learned policies seem to suggest this is not the case. In fact, the unreliability of the policy in the later stages of training seems to be related to the agents tendency to spread more around the map, even if this sometimes means ignoring the safe zone. This may be because agents didn't learn to avoid hurting each other, as in the free--for--all variants 2 and 3, and they instead have learned that standing near other agents is dangerous. In fact, theoretically, since the damage done by an agent's attack is much more powerful than the damage done by the safe zone, it is more important to avoid other agents in the short term with respect to staying inside the zone. Another reason, besides chance, that may have prevented agents from avoiding to attack opponents is the teaming mode. For team--based variants, since the reward is shared between teammates, the death of one teammate does not directly affect the reward, and thus there is more tolerance for sloppy teammates. This is the opposite of what happens, for example, in \citeauthor{hideseek}, in which the ``hide and seek'' reward scheme directly punishes hiders that act poorly.

The only two emergent behaviours shown by these experiments are thus:
\begin{enumerate}
    \item ``pacifist'' agents, in free--for--all variants 2 and 3; and
    \item agents spreading around the map to avoid being killed, in most of the other variants.
\end{enumerate}
In both cases, this means that agents learned to actively avoid combat interaction with the other agents. In the second case, this also suggests that training these configurations for more environment steps may not give rise to any other interesting emergent interactions. In the first case, instead, since agents stop attacking but don't actively run away from each other, combat interaction may still arise if trained more on the last phases of the episodes, in which fighting for a spot in the safe zone is crucial.

The reported results and the above considerations also highlight the following flaws in the environment variants considered:
\begin{enumerate}
    \item Agents don't ever hold onto heal items, and never give them to another agent; instead, they always consume them as soon as they are collected. This is because there is no clear advantage in not doing so: using a heal at full health has the same effect of using a heal at 1 health.
    \item Team rewards are such that a sloppy teammate does not affect the team reward, at least in the early phases of each episode. This is because the only advantage in all teammates surviving is the possibility of teaming up in combat against an opponent, or exchanging heals. Since the latter is unlikely due to flaw 1 and combat interactions are very sparse, this reward scheme produces teams with less consistent agents than the free--for--all configuration.
    \item The magnitude of the reward for kills and deaths does not result in high combat interaction, and makes high--competitiveness variants more similar to those with medium competitiveness. In fact, since the episodes in both configurations end when only one agent or team is alive, it is better for agents to not engage in combat early in the episode, because surviving the combat may bring a much higher reward with respect to the kill reward.
    \item The dynamic controls for agent motion may be too limiting to allow for precise movements, resulting in less consistent performance overall.
\end{enumerate}

\newpage

\section{Conclusions and Future Work}


The two main contributions of this work are:
\begin{itemize}
    \item developing a modular and efficient framework for 2D multi--agent environments, and implementing concrete, survival--based environments on top of it;
    \item conducting experiments on the developed variants, obtaining some basic emergent behaviours, and identifying concrete directions for the improvement of the tested environment variants.
\end{itemize}

Future work should address the flaws listed above before focusing on further experiments. Flaw 1 can be easily solved by introducing a maximum value for the health. Flaw 2 should be addressed by coming up with a different reward scheme for team--based variants, so that sloppy teammates are less tolerated, or by adding mechanics to the environment that promote team play. In this regard, solving flaw 1 may help by giving more value to inventory--based mechanics, such as sharing items or killing opponents to collect the items they were holding. Flaw 3 can be solved by either dismissing the kill for rewards, focusing more on indirectly competitive variants, or by simply increasing the reward given for kills. 

Considering that the training scale at which the above experiments were conducted is significantly smaller than related work, such as \citeauthor{hideseek}, it may also be beneficial for future work to focus on fewer environment variants. This would allow to allocate a higher number of training steps to each variant, avoiding the need for more computing resources.


    \newpage
\bibliographystyle{apalike}
\bibliography{bibliography}

    \appendix
\section{Environment Implementation} \label{s:impl}

\begin{table}[b]
    \centering
    \begin{tabular}{|l|c|}
        \hline
        \textbf{Features} & \textbf{Seconds per step ($\mu \pm \sigma$)} \\
        \hline
        5 agents (no cameras) & $4.77\cdot10^{-4} \pm 6.70\cdot10^{-5}$ \\
        5 agents (no cameras), 10 heals & $4.94\cdot10^{-4} \pm 8.50\cdot10^{-5}$ \\
        5 agents (no cameras) 10 heals, 10 boxes & $5.07\cdot10^{-4} \pm 7.09\cdot10^{-5}$ \\
        5 agents & $5.72\cdot10^{-4} \pm 9.36\cdot10^{-5}$ \\
        5 agents, 10 heals, 10 boxes & $7.24\cdot10^{-4} \pm 9.21\cdot10^{-5}$ \\
        10 agents, 10 heals, 20 boxes & $1.64\cdot10^{-3} \pm 1.47\cdot10^{-4}$ \\
        10 agents (no cameras), 10 heals, 20 boxes & $8.03\cdot10^{-4} \pm 1.47\cdot10^{-4}$ \\
        \hline
    \end{tabular}
    \caption{Computational performance tests on the environment step function. Each result reports the mean and standard deviation of the time taken by each step of the simulation object with the given features. The tests were run on a Macbook Pro '15 laptop, with an Intel Mobile Core i7 ``Broadwell'' (I7-5557U) CPU.}
    \label{tab:perf}
\end{table}

The environment is implemented in the Python programming language. It features a Gym \cite{gym} interface, slightly modified to account for multiple agents. Moreover, it also supports the VecEnv interface to use with the Stable Baslines 3 library \cite{sb3}, by making the multi--agent environment appear as a vectorized single--agent environment. With the goal of making the implementation modular and extensible, an abstract framework has been developed, discussed in section \ref{ss:framework}. The Python bindings for the Box2D library were used as the underyling physics simulator, since Box2D provides an efficient and compact interface. The optional rendering of the environment state uses PyGame \cite{pygame} as the backend.


\subsection{Framework} \label{ss:framework}

The framework of the environment features a core object, the \emph{simulation}, which can be extended by adding \emph{groups} of bodies, each containing of one or more \emph{modules}. The simulation object simply wraps an underlying physics engine to support the addition of groups. Each group represents a collection of bodies which share the same set of modules, and thus have similar behaviour. The specific semantics are then implemented in the modules themselves, which can, and often will, maintain body--specific data. This architecture allows to retain efficiency by grouping bodies with similar behaviour together, while also not taking away any extensibility or modularity. Moreover, modules are designed as collections of callbacks which are automatically called by the group and simulation objects, so that new modules can be added without modifying any existing part of the code. The specifics of the simulation, group and module interfaces are described in the remainder of this section; core modules that are part of the framework are described in section \ref{ss:coremods}. The external modules implementing the semantics specifically used in the experiments of this work are detailed in later sections.

A \emph{simulation} only features two operations: resetting and stepping. The former resets the state of the physics engine and of all the groups, while the latter performs one time step of simulation. Depending on the simulation instance and the physics engine, a single time step of the simulation may correspond to several time steps of the physics simulator. The state of a simulation object is the state of the underlying physics simulator, plus a list of references to its groups.

A \emph{group} object provides the following operations:
\begin{itemize}
    \item \texttt{spawn}, which spawns new bodies in the group;
    \item \texttt{despawn}, which makes existing bodies in the group despawn;
    \item \texttt{get\_module}, which returns an instance of a given module class that is in the group's modules.
\end{itemize}
The state of a group object will thus consist of a list of modules, a list of live bodies and possibly a reference to a simulation object that the group belongs in.

A \emph{module} object implements zero or more of the following callbacks:
\begin{itemize}
    \item \texttt{post\_reset}, called upon the resetting of the simulation object;
    \item \texttt{pre\_step}, called before executing each simulation step;
    \item \texttt{post\_step}, called after executing each simulation step;
    \item \texttt{post\_spawn}, called after one or more bodies have spawned in the module's group;
    \item \texttt{pre\_despawn}, called when one or more bodies in the module's group are about to despawn.
\end{itemize}
The state of a module object is completely left to its implementation.

\subsection{Built--in Modules} \label{ss:coremods}

This section overviews general--purpose modules provided in by the environment framework itself. These mainly implement applying forces to bodies, observing the world in various ways, and tracking useful information.

The \texttt{IndexBodies} module assigns unique integer indices to all bodies that spawn in its group. It does so by keeping a dynamic array of references to body objects, so that the index of each body is its index in the array. When a body despawns, its entry in the array is not deleted to keep the indices unique, and is instead replaced with a null reference. This makes it also useful to quickly test whether the body corresponding to a certain index has despawned or not.

The \texttt{TrackDeaths} module keeps track, for each simulation each step, of all the bodies in the group that have despawned. The \texttt{LogDeaths} module is similar, but also prints a message to the console when a body in the group despawns.

The \texttt{Cameras} module attaches ``cameras'' to all bodies in the group. Each body in the group is associated with a vision cone originating from its position. For each observing body in the group, the module stores, at a given time step, all the bodies in the world that are visible to the observer. To be visible, a body has to be inside the vision cone of the observer, and its line of sight to the observing body must not be obstructed by other bodies.


The \texttt{DynamicMotors} module provides dynamic control of the bodies in the group. At each time step, a specific linear and angular forces can be applied to each body in the group. Note that the controls are specific to each body.

\subsection{General Purpose Modules}


Several external Modules implement the environment semantics specific to this work. Some of them are only used in a single body group, while others serve more general purposes are used in multiple groups. The three main modules that are used in multiple groups are \texttt{ResetSpawns}, \texttt{Health} and \texttt{Item}. The main modules specific to the agent body group are \texttt{SafeZone}, \texttt{Inventory} and related modules, \texttt{Melee} and \texttt{TwoTeams}. Item body groups use different implementations of the abstract \texttt{Item} module. The heal items body group mostly uses the \texttt{Heal} item module. The boxes body group mainly uses the \texttt{RandomizedBoxShapes}, and \texttt{OwnedObject} modules, while the box items body group is based on the \texttt{OwnedObjectItem} item module. The remained of this section describes in more detail the semantics that each module adds to the core environment.

\texttt{ResetSpawns} simply implements a \texttt{post\_reset} callback that spawns bodies in the group using a given spawner object. The spawner object simply provides the placements for the bodies to be spawned. Since it can be shared between different \texttt{ResetSpawns} modules, possibly in different body groups, the logic that determines the spawning placements can take into account all the bodies at once, even if they belong to different groups. The specific spawner used here simply takes a rectangular region, divides it into cells, and assigns each body a random unused cell to spawn.

The \texttt{Health} module is used to assign an integer health value to all the bodies in the group, and provides an interface to other modules to damage and heal those bodies. This interface also supports tracking the causes of damage and healing, adding filters that make a body immune to some causes, and registering an \texttt{on\_death} callback that also provides the cause of death of a body.

The abstract \texttt{Item} module simply defines the interface that all modules that implement items should follow. This is so that, for example, the modules that implement inventories or other item--related mechanics can disregard the specific type of items involved. A module is an item module if it implements the \texttt{use} operation, called when an item is consumed for use. The abstract \texttt{Item} module also provides the concrete \texttt{drop} operation, which spawns an item in its body group with a given object as the item--specific data, which is is later passed to \texttt{use} when the item is consumed.

\subsection{Agent Specific Modules}

The \texttt{SafeZone} module implements the safe zone mechanic described above. It does so by maintaining as state simple integer counters that keep track of which phase the zone is in, and the remaining steps to the next phase. Damage is dealt to the bodies in the group assuming the group contains a \texttt{Health} module.

The agent's inventory is implemented in the \texttt{Inventory} module. For each body in the group, this module keeps a fixed--length array of item slots, and provides the operations:
\begin{itemize}
    \item \texttt{pickup}, which makes a body pick up an item on the ground;
    \item \texttt{use}, which attempts to make a body use the item in the last slot of its inventory;
    \item \texttt{drop}, which attempts to make a body drop on the ground the item in the last slot of its inventory;
    \item \texttt{drop\_all}, which makes a body drop all the items in its inventory on the ground;
    \item \texttt{give}, which attempts to make a body give the item in its last slot to another body that has an inventory.
\end{itemize}
The \texttt{Inventory} does not automatically perform these actions, but rather makes them available to other modules. In fact, to implement the inventory--related agent mechanics, these operations were used in the modules \texttt{AutoPickup}, \texttt{UseLast}, \texttt{GiveLast} and \texttt{DeathDrop}. \texttt{AutoPickup} makes all agents in its group pickup items when the agent touches the item object. \texttt{UseLast} and \texttt{GiveLast} provide the controls to make an agent use or give the item in its last slot of the inventory. Finally, \texttt{DeathDrop} makes the bodies in its group drop all items in their inventory when they are about to despawn, slightly scattered in a circle around their last position.

The attack action of the agents is implemented in the \texttt{Melee} module. This module provides one binary control for each agent. At each step, if a control for an agent is on, the module tries to find a body with an \texttt{Health} module that is in range. If it finds a suitable target, this is damaged, marking the attacking body as the cause of damage.

Subdivision of the agents in teams is implemented in the \texttt{Teams} module. \texttt{Teams} keeps track of all members of the teams, and provides operation to get the teammates or the opponents of an agent. Additionally, it also provides an operation to test whether all the agents in the team are dead. Besides providing team bookkeeping, it also overrides the cause of damage of \texttt{Melee} attacks to the team, and makes agents on the same team immune to attacks by teammates.


\subsection{Heals and Boxes Modules}

The only heal--item--specific module is \texttt{Heal}. This is a concrete \texttt{Item} module, for which the \texttt{use} operation simply increases the health of the user by an amount dependent on how the \texttt{Heal} module was initialized. 

Modules specific to the breakable boxes are mainly the ones that implement the logic of turning broken objects into items which can be used to respawn them. The \texttt{Object} and \texttt{ObjectItem} modules are responsible for producing this behaviour in the two body groups in which these entities are split: intact objects will be in the group that has the \texttt{Object} module, while broken objects will be in the group with the \texttt{ObjectItem} module. These two modules maintain references to each other and keep the data relative to the shape and other properties of the objects. The \texttt{Object} module used this data to spawn an appropriate item when an object body in its group is about to despawn. Conversely, the \texttt{ObjectItem} spawns an object in the group of the \texttt{Object} module, using the recorded data, when an agents collects and uses that item. The \texttt{OwnedObject} and \texttt{OwnedObjectItem} implement a variant of objects that only allows the entity that first broke an object to break it again when it is placed. Ownership can be attributed to single bodies or to other entities such as teams.


\subsection{Rendering Framework}

An optional rendering framework was also implemented, using the PyGame library \cite{pygame} as a backend. The design mimics the extensibility and modularity features described above, with the main difference being that there is no need to model body groups explicitly. A single rendering module is called a \emph{View}, while the object that interfaces Views with the rendering backend is the \emph{Canvas}. Each View implements a single drawing callback, which is used by the Canvas it belongs to. A Canvas object, on the other hand, maintains state relative to the rendering backend and a list of Views. It also implements rendering and drawing operations. More specifically, rendering operations are clearing and rendering, and can be called by the Canvas user each time a frame image is needed. Clearing simply clears the screen with a background color. Rendering calls all the draw callbacks of each registered View, and then calls the backend to actually display the rendered frame on screen. Drawing operations, on the other hand, are not meant to be called by the Canvas user, but rather are available for use inside each View's implementation. They provide drawing of basic shapes such as circles, segments and polygons, as well as the drawing of bodies in the world.

\end{document}